\documentclass[10pt,twocolumn,letterpaper]{article}

\usepackage{cvpr}
\usepackage{times}
\usepackage{epsfig}
\usepackage{graphicx}
\usepackage{amsmath}
\usepackage{amssymb}
\graphicspath{ {advphotos/} }
\usepackage{subfig}

\usepackage[pagebackref=true,breaklinks=true,letterpaper=true,colorlinks,bookmarks=false]{hyperref}

\cvprfinalcopy 

\def\cvprPaperID{****} 

\ifcvprfinal\fi
\begin{document}

\title{Traits \& Transferability of Adversarial Examples against Instance Segmentation \& Object Detection}

\author{Raghav Gurbaxani \hspace{1cm} Shivank Mishra\\
University of Illinois at Urbana Champaign\\
{\tt\small {\{raghavg3,smishr25\}} @illinois.edu}
\and
\\
{\tt\small  }
}

\maketitle

\begin{abstract}
Despite the recent advancements in deploying neural networks for image classification, it has been found that adversarial examples are able to fool these models leading them to misclassify the images. Since these models are now being widely deployed, we provide an insight on the threat of these adversarial examples by evaluating their characteristics and transferability to more complex models that utilize Image Classification as a subtask.

We demonstrate the ineffectiveness of adversarial examples when applied to Instance Segmentation \& Object Detection models. We show that this ineffectiveness arises from the inability of adversarial examples to withstand transformations such as scaling or a change in lighting conditions. Moreover, we show that there exists a small threshold below which the adversarial property is retained while applying these input transformations. 

Additionally, these attacks demonstrate weak cross-network transferability across neural network architectures, \emph{e.g.} VGG16 and ResNet50, however, the attack may fool both the networks if passed sequentially through networks during its formation.

The lack of scalability and transferability challenges the question of how adversarial images would be effective in the real world.
\end{abstract}
\section{Introduction}

With recent advancements in Computer Vision, Neural Networks have been able to achieve state-of-the-art results in Image Recognition and have been able to outperform human-level performance. The success of Convolutional Neural Networks \cite{Alexnet} ,\cite{VGG},\cite{Resnet}, \cite{Inception} on the Imagnet \cite{ImageNet} dataset (for the Large-Scale Image Recognition Challenge) has propelled neural networks to achieve significant results in various visual recognition tasks. Consequently, they are now widely deployed at an unprecedented scale across a variety of applications such as Robotics, Drones, Self-Driving Cars, and Surveillance.

On the other hand, the discovery of adversarial examples \cite{Intrigue}, \cite{fool}, \cite{FGSM} has threatened these achievements by fooling neural networks to misclassify the image by introducing almost visually imperceptible perturbations. Adversarial examples  have been able to fool the state of the Image Classification models with a high confidence score. as shown in Figure1 Moreover, \cite{Kurakin} showed that adversarial can exist in the physical world. To put this in perspective, an adversarial image of a stop sign on road could be misclassified as a green light - thus raising a huge security threat.

As Image classification based models are now being widely deployed across the industry, it is consequential to study of the effects of these adversarial perturbations. Keeping in mind the harmful ramifications, in this paper we study the properties of adversarial examples in real world conditions and evaluate their effectiveness across different neural networks architectures.

The advancements of neural networks Classification tasks has propelled research in developing more complex algorithms such as Object Detection \cite{fasterRCNN}, \cite{RetinaNet} which use Image Classifiers as a subtask and to identify all the objects in an image, as well as localize their position in the image. Moreover, Instance Segmentation \cite{MaskRCNN} further extends Object Detection by adding a segmentation pipeline to it, thus allowing it to not only detect objects but also segment each pixel on those objects leading to a more fine-grained classification.
With the advent of these complex models, which use image classification as a sub-task and taking into account the susceptibility of Image classifiers against adversarial images, it becomes important to study whether the adversarial image is able to fool these advanced pipelines as well. With the advent of these complex models and taking into account the threat of adversarial attacks, it becomes necessary to examine whether these attacks are able to mislead these complex advanced model pipelines as well.

To study the various properties of adversarial examples we use the FGSM{\cite{FGSM}}, DeepFool\cite{DeepFool}, and Carlini \& Wagner L2 \cite{cnw} attacks in a targeted and untargeted setting. To simulate the real world we a variety of scaling, rotation and lighting conditions and demonstrate that these attacks do not retain their adversarial property under these situations. Interestingly, we discovered that in case of these transformations, there exists a minimum threshold below which the adversarial property is retained and is lost in the latter case. We then study the transferability of these attacks to various network architectures on a Cross Network (same dataset, different architectures) and a Cross Task Transfer (different dataset, similar architecture). We notice that adversarial attack demonstrate weak transferability across different networks (VGG, ResNet,etc.) on the ImageNet dataset. Moreover, these transformations deem these attacks ineffective against Cross Task models such as Object Detection \& Instance Segmentation models– which involve cropping objects in the image and classifying them individually to their respective classes. We further chose a third proxy task of Image Captioning and show that adversarial examples are not transferable unless the setup is as maintained by \cite{Show}. 

We further extend the study conducted by \cite{noneed} to Instance Segmentation and Captioning models and study the scalability of attacks in the real world.

\begin{figure}[h!]
\begin{center}
    \subfloat[Box Turtle]{{\includegraphics[width=2.75cm]{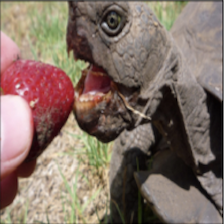}}}
    \qquad
    \subfloat[African Chameleon]{{\includegraphics[width=2.75cm]{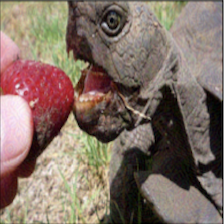}}}
\caption{Example of adversarial misclassification}
\end{center}
\end{figure}
\section{Adversarial Attacks}
\subsection{Fast Gradient Sign Method}

The Fast Gradient Sign Method, as introduced by \cite{FGSM}  exploits the linearity of the neural network models in a high dimensional space. We used the targeted FGSM attack – where the attacker specifies the adversarial image class, to which is should be misclassified.

Consider \emph{x} as the original image, \emph{y} the corresponding label of \emph{x}, $\theta$ the parameters of the network and L($\theta$,\emph{x},\emph{y}) the loss function used to train the network. 

\textbf{\begin{center}$\eta = \epsilon sign(\nabla_xL(\theta,\emph{x},\emph{y}))$\end{center}}

The attack, essentially an optimization problem, tries to increase the loss of the classifier. It takes the derivative of the loss function in the x-direction - $\nabla_xL(\theta$,\emph{x},\emph{y}) and the sign of the derivative to determine the direction of the pixel change. The sign term ensures that the loss is maximized.  Moreover, this is multiplied by a small constant $\epsilon$, which ensures that the perturbation doesn’t go too far from the original image and restricts its norm that gives us the perturbation $\eta$.

This perturbation can now be added to the original image \emph{x} to result in the adversarial image, capable of fooling the classifier.

$x_{adv}$ = \emph{x} + $\eta$
For our experiments, we used the Cleverhans \cite{cleverhans} library to generate the adversarial images on pre-trained Imagenet models with y\_target set to class 'tennis ball' and $\epsilon$ set to 0.3.

\subsection{DeepFool}
The perturbations generated by Deepfool are smaller than the perturbations compared to FGSM in terms of the norm but have similar effectiveness in terms of fooling ratios as shown in Figure2. It also reduces the intensity of the perturbations and was used in the untargeted attacking case – where the attacker can only specify that the adversarial image is classified differently than the original image.
 
This method  \cite{DeepFool} was used to iteratively minimize the distortion which leads the image to switch classes by projecting the input image to the closest separating hyperplane (considering each class of classifier is separated by a hyperplane) and by linearizing the decision boundaries in which the image confines.

For our experiments, we used the Foolbox\cite{foolbox} library to generate the adversarial images on pre-trained Imagenet models with 50 iterations and $\epsilon$ set to 0.3.

\begin{figure}[bhp]
\begin{center}
    \subfloat[Ballplayer]{{\includegraphics[width=2.75cm]{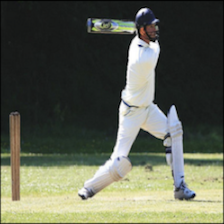} }}
    \qquad
    \subfloat[Chain Mail]{{\includegraphics[width=2.75cm]{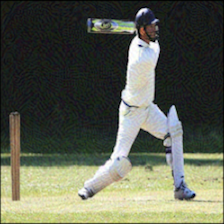}}}
\caption{Example of Original \& Adversarial Image Generated from DeepFool Attack}
\end{center}
\end{figure}
\subsection{Carlini \& Wagner L2}
The Carlini \& Wagner L2 (C\&W L2) attack \cite{cnw} was introduced to deal with the defensive distillation strategy, which made the perturbations quasi-imperceptible. Although the distortions produced by C\&W L2 are smaller than the previously mentioned attacks, it is also a slower attack than FGSM. These attacks limit the perturbation by restricting their L2 norm.
\section{Experimental Setup}

In this section we describe the datasets, the deep neural network models, adversarial attacks methods, the evaluation metrics and the deep learning frameworks that we used to carry out our experiments.

\subsection{Datasets}

In order to generate the adversarial examples, we used the pre-trained models on the ImageNet dataset. Then, to evaluate the generated adversarial examples on the Object detectors and Instance Segmentation models, we used models pre-trained on the MSCOCO\cite{mscoco} dataset.

For Image Captioning, we used models pre-trained on the MSCOCO dataset which comprises of 82783 training images, and 40504 and 40775 validation and test images respectively with each image corresponding to 5 captions each.

\subsection{Models}

For object detection we used the Faster RCNN model and the RetinaNet models, each using the ResNet 50 pipeline. We also used the Mask RCNN Instance Segmentation module, which utilizes a Feature Pyramid Network and a ResNet 101 CNN backbone . Lastly, We used the Image Captioning pipeline from Show and Tell based on a VGG 16 CNN.
The adversarial examples based on the three attacks were generated using VGG16, ResNet 50 and Resnet 101 models pre-trained on the ImageNet dataset. 

\subsection{Attacks}

We employed the FGSM, DeepFool and C\&W L2 using pre-trained models on ImageNet. We have mentioned the hyperparameters that we used for the respective attacks in Section 2. We used the FGSM and C\&W L2 attacks in a targeted setting whereas the DeepFool attack in an untargeted setting. 

\subsection{Metrics}

We used Intersection over Union (IoU) as the primary metric while evaluating Instance Segmentation and Object Detection models. For Image Classification, we utilized accuracy as a metric. For Image Captioning, we utilized the BLEU score.

\subsection{Deep Learning Frameworks}

We used the pretrained models on ImageNet using the Keras Library which uses Tensorflow backend. In order to generate the FGSM and C\&W L2 attacks, we used the novel Cleverhans library released by Goodfellow et. al. For the DeepFool attack we employed the Foolbox library.

\section{Characteristics of Adversarial Attacks}

To gain complete understanding of Adversarial examples, we have surveyed some of their properties in this section by subjecting them to several scaling, rotation, and lighting conditions. 

\subsection{Cropping}

For the latter question, we were able to notice that when the image was cropped from the original size of 224 x 224 to 224 x 223 (1 row cropped), it retained its adversarial property. However, it suffered a significant drop in adversarial confidence score as shown in Figure 3.
On further cropping, it loses its attacking capability and the classification switches back to the original class (Labrador retriever in this case). This would indicate towards the idea that there exists a global pattern for each adversarial image that must be preserved for it to be able to attack classifiers.

\begin{figure}[h!]
    \subfloat[224 x 224  ]{{\includegraphics[width=0.25\columnwidth]{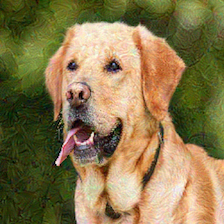} }}
    \qquad
    \subfloat[223 x 224]{{\includegraphics[width=0.25\columnwidth]{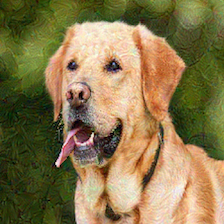} }}
    \qquad
    \subfloat[223 x 222]{{\includegraphics[width=0.25\columnwidth]{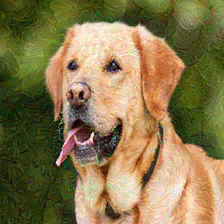} }}
    
\texttt{}
Predictions:- \\
\phantom{x}\hspace{1.8ex}a)  $ tennis\_ ball (75.65\%$)\\
\phantom{x}\hspace{1.9ex}b) $ tennis\_ ball (50.25\%)$\\
\phantom{x}\hspace{1.9ex}c) $ golden\_ retriever (56.49\%)$

    \caption{Demonstrates the a) Adversarial Image b) Adversarial Image after 1 row cropped c) Cropped further - cropping beyong a certain limit causes the image to lose it's adversarial class). }%
    \label{fig:example}%
\end{figure}
\subsection{Magnification}
Magnification can be considered as a consequence of cropping as it leads to enlargement of the cropped area and thus also leads to a loss of attacking ability for the adversarial image as shown in Figure 4. 

\begin{figure}[h!]
	\hspace{5.8ex}
    \subfloat[Adversarial ]{{\includegraphics[width=2.75cm]{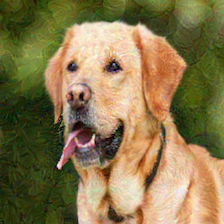} }}
    \qquad
    \subfloat[Magnified Adversarial]{{\includegraphics[width=2.75cm]{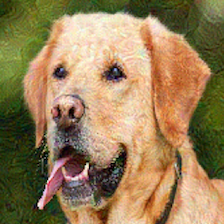}  }}
\texttt{}
\vspace{0.01cm}\\
Predictions:- \\
\phantom{x}\hspace{1.8ex}a)  tennis\_ball$(75.65\%)$\\
\phantom{x}\hspace{1.9ex}b)  Labrador retriever $(52.66\%)$\\

    \caption{The effect of Magnification - In Fig  b) magnification  magnification causes adversarial image to lose it's attacking class }%
    \label{fig:example}%
\end{figure}

\subsection{Rotation}
Similar to cropping and magnification, adversarial images were not able to maintain their attacking property on being rotated. We were curious to determine a threshold for the susceptibility to rotation at which the image would lose it's attacking ability. However, we found that the images lose their attacking capability even with a 1-degree rotation (in both clockwise and anti-clockwise directions) as shown in Figure 5.
\begin{figure}[h!]
	\hspace{5.8ex}
    \subfloat[Adversarial Image]{{\includegraphics[width=2.75cm]{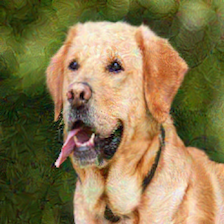} }}
    \qquad
    \subfloat[Rotated Adversarial Image]{{\includegraphics[width=2.75cm]{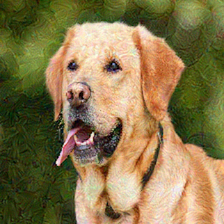} }}
\texttt{}
\vspace{0.01cm}\\
Predictions:- \\
            \phantom{x}\hspace{1.9ex}a)tennis\_ball $(75.65\%)$\\
      \phantom{x}\hspace{1.9ex}b)  Labrador\_ retriever $(45.20\%)$\\
    \caption{The effect of Rotation - In Fig  b) rotation (clockwise \& anti- clockwise)  causes adversarial image to lose it's attacking class }%
    \label{fig:example}%
\end{figure}
\texttt{\noindent}

\subsection{Change in brightness}
In order to consider how adversarial images would perform in the real world under myriad of lighting conditions, we subjected these images to change in exposure. Our experiments show that adversarial examples are robust to change in exposure and retain their attacking capability until a certain threshold. As demonstrated in Figure 6, a 50\% increase in exposure in the adversarial image retains the adversarial class, but a 100\% increase in exposure causes it to return to its original class.

\begin{figure}[h!]
    \subfloat[No change in brightness ]{{\includegraphics[width=0.25\columnwidth]{Picture11.png} }}
    \qquad
    \subfloat[50\%  increase]{{\includegraphics[width=0.25\columnwidth]{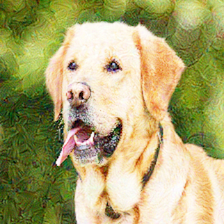} }}
    \qquad
    \subfloat[100\%  increase]{{\includegraphics[width=0.25\columnwidth]{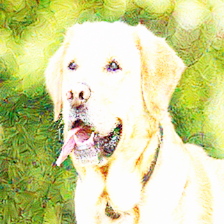} }}
    
\texttt{}
\vspace{0.01cm}\\
Predictions:- \\
\phantom{x}\hspace{1.9ex}a)tennis\_ball $(75.65\%)$\\
\phantom{x}\hspace{1.9ex}b) tennis\_ball $(76.42\%)$\\
\phantom{x}\hspace{1.9ex}c) golden\_retriever $(43.11\%)$
    \caption{The effect of Brightness - In Fig   b) adversarial image with 50\%  increase remains adversarial, however in Fig. c)100\% increase reverts to original class}%
    \label{fig:example}%
\end{figure}

\section{Transferability of Attacks across Architectures}

We evaluate the transferability of of adversarial images across -
\newline1. Cross-Network Transfer: Using different architectures on a common dataset
\newline2. Cross-Task Transfer: Using different architectures on different datasets

\subsection{Cross-Network Transferability}

In this section, we investigate how well do adversarial images generalize over different neural network architectures while keeping the dataset and the task constant. In order to do this task, we use a computed adversarial image from one model and use it to attack different model architecture. We tried the following experiment in a targeted attack setting on various models on the ImageNet dataset.

As demonstrated in  Figure 7, we generated the adversarial image using a ResNet 50 based classifier that was able to fool the ResNet 50 pipeline with high confidence. However, when we transferred this adversarial image for evaluation on a different architecture such as the VGG 16, it reverted to its original (non-adversarial) class. 

We were able to achieve similar results while performing the experiments vice-versa and with other models such as ResNet 101 and Inception V3. This indicates that the adversarial examples are weakly transferable across neural network architectures on the same task.

\begin{figure}[h]
	\hspace{5.8ex}
\subfloat[Original Image]{{\includegraphics[width=2.75cm]{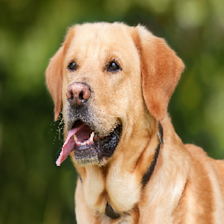} }}
    \qquad
    \subfloat[Adversarial Image]{{\includegraphics[width=2.75cm]{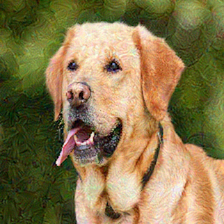} }}
\texttt{}
\vspace{0.01cm}\\
Predictions:- \\
\phantom{x}\hspace{1.9ex}a) Resnet 50 - Labrador\_retriever (58.70 $\%$)\\
\phantom{x}\hspace{1.9ex}b1) Resnet 50 - tennis\_ball (75.65$\%$)\\
\phantom{x}\hspace{1.9ex}b2) VGG 16 - golden\_retriever (42.38$\%$)

    \caption{Cross Network Transferability: Fig a) shows the classification for original image on ResNet 50, for Fig b1) shows adversarial fooling on ResNet 50, Fig b2) however when adversarial image is passed to VGG- fails to transfer and reverts to original class}%
    \label{fig:example}%
\end{figure}

\subsection{Cross-Task Transferability}
In order to evaluate Cross Task Transferability, we used adversarial images generated through pre-trained models on the ImageNet dataset and evaluate their transferability across three tasks that have a different model pipeline and were trained on a different dataset.

\subsubsection{Object Detection}

We first evaluate our cross-task task transferability on object detection models Faster RCNN and RetinaNet which used ResNet 50 based Image Classification model as a subroutine.

Object detection models form bounding boxes over the objects and then pass it to the classification model - it effectively translates to cropping the objects under the bounding box leading to classification under different resolutions. 

As seen in section 4.1, the cropping operation renders the adversarial image ineffective for an attack, thus detectors are not affected by these adversarial images.

The Figure 8 below demonstrates that the adversarial attack has no visible impact on the results of the detection pipeline.
\begin{figure}[h!]
    \centering
    \subfloat[Original Detection]{{\includegraphics[width=2.75cm]{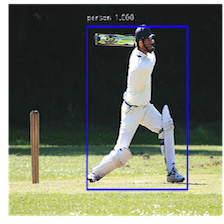} }}
    \qquad
    \subfloat[Adversarial Detection]{{\includegraphics[width=2.75cm]{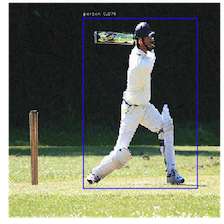} }}
    \caption{Object detection, with and without adversarial attack. }%
    \label{fig:example}%
\end{figure}
\subsubsection{Instance Segmentation}

In order to extend the results obtained on object detectors in the previous section, we evaluate the attacks on the Instance Segmentation models. 
Despite passing the adversarial image to the models, there is no discernible difference in the results of the Mask RCNN pipelines between the original and the adversarial images as shown in Figure 9.
The instance segmentation models utilize object detection as a subroutine and thus adversarial attack does not cause a change in result due to the susceptibility to the cropping operation during formation of the bounding boxes.

\begin{figure}[h!]
    \centering
    \subfloat[Original Inst. Segmentation]{{\includegraphics[width=2.75cm]{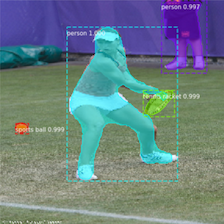} }}
    \qquad
    \subfloat[Adversarial Inst. Segmentation]{{\includegraphics[width=2.75cm]{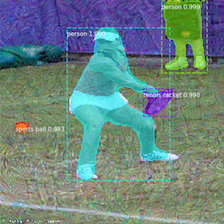} }}
    \caption{Instance Segmentation: with and without Adversarial attack}%
    \label{fig:example}%
\end{figure}

\subsubsection{Image Captioning}
Lastly, we assess the transferability of adversarial attacks to the Image Captioning models. The intent was to recognize how the adversarial perturbations affect an RNN based model and how they alter the captions and their semantic properties. For the experiment, we used a captioning model based on a CNN-LSTM (encoder-decoder) pipeline,pre-trained on the  MSCOCO dataset. The model used with k beam search inference technique keeping k =1. Following are some of the results obtained.

  


\begin{figure}[h!]
	\hspace{5.8ex}
    \subfloat[Original Image]{{\includegraphics[width=2.75cm]{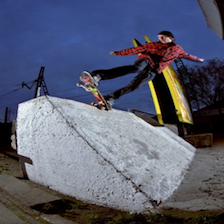} }}
    \qquad
    \subfloat[Adversarial Image]{{\includegraphics[width=2.75cm]{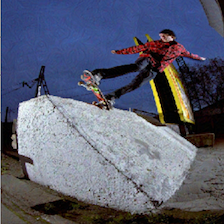} }}

\texttt{}
\vspace{0.01cm}\\
{Generated Captions:}\\
a) A man flying through the air while riding a snowboard. (p=0.021026)\\
b) A person jumping a snow board in the air (p=0.007723)\\
    \caption{Demonstrates a) original and b) adversarial image and respective captions generated.} 
    \label{fig:example}%
\end{figure}

  


  
  
\begin{figure}[h!]
	\hspace{5.8ex}
    \subfloat[Original Image]{{\includegraphics[width=2.75cm]{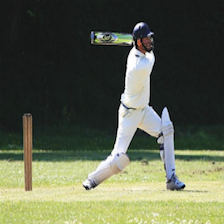} }}
    \qquad
    \subfloat[Adversarial Image]{{\includegraphics[width=2.75cm]{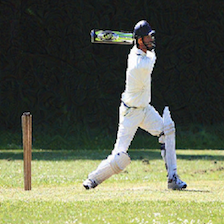} }}

\texttt{}
\vspace{0.01cm}\\
{Generated Captions:}\\
a) A baseball player throwing a baseball on a field. (p=0.001463)\\
b) A baseball player swinging a bat at a ball (p=0.001510)\\
    \caption{Demonstrates a) original and b) adversarial image and respective captions generated.}   
    \label{fig:example}%
\end{figure}

It is apparent in Figures 10 \& 11 that the introduction of adversarial perturbations does not fool the captioning pipeline as the captions obtained by adversarial and original images are identical. 

\subsection{Attacks Capable of Fooling Multiple Networks}
As an extension of the study described in section 5.1, we examined whether adversarial examples could fool multiple network architectures simultaneously.

We passed the adversarial image generated from one model architecture as the input for generation to the next model architecture. In several cases, we discovered that the resulting adversarial image was capable of fooling both the networks architectures.

In the image below, we carried out the aforementioned process by generating the adversarial image using a ResNet 50 pipeline. We then transferred it for generation using a VGG16 model and found that the resulting image was able to fool both the classification models (tennis\_ball (99.49\%) against vgg16 and tennis\_ball (67.52\%) against resnet 50).

However, the resulting image also experienced more visual distortion. This experiment would also suggest that it might possible to devise an adversarial example, capable of fooling all the major neural network architectures.



\section{Future Work}
 In this section, we suggest some directions for future researchers to consider - 

\subsection{Change Loss Function in Object Detection \& Instance Segmentation}
As demonstrated by Chen \emph{et. al} that adversarial images can be transferred to Image Captioning models by altering the loss function of the model. Whether similar results can be achieved on Object Detectors and Instance Segmentation models remains to be seen. 

\subsection{Effect of adversarial examples on Generative models}
So far, adversarial images have been widely deployed against discriminative models. However, the effects of adversarial examples on generative models such as Generative Adversarial Networks and Autoencoders needs to be researched. Xiao et al. \cite{xiao} and \cite{Samangouei} showed how GANs can be employed for adversarial attack and defense respectively. However detailed studies on VAEs and GANs for adversarial images have yet to be conducted. One question that needs to be answered is whether adversarial examples can be generated on a large scale using generative models without a loss of clarity and attacking ability.

\subsection{Effect of precision on adversarial examples}

Most of the existing research on adversarial examples has been carried out on 32-bit float models. As most vision models in real-life applications are being deployed using lower precision of these neural networks weights such as 16-bit float or 8-bit integer. It would be intriguing to examine the effects of adversarial examples with lower precision going down till binary.

\subsection{Comparison of network architectures instead of fooling capability}
Another possible direction would the effect of adversarial examples against various neural network architectures. It is important to see how various attacks fare against different architectures; which architectures are able to defend against these attacks better.

\section{Conclusion}
 
In this paper, we showed that adversarial examples are weakly transferable across neural network architectures – especially in more complex vision tasks. Using the FGSM, DeepFool and C\&W L2 attacks, we evaluated the cross-network transferability of adversarial images across network architectures on the ImageNet dataset. We extended our study whether these attacks are transferable to more complex Computer Vision tasks such as Instance Segmentation and Object Detection

As in the real world, adversarial images may be encountered in a variety of scale and lighting conditions we subjected these images to various input transformations such as cropping, magnification, rotation, and change in lighting conditions. We found that there exists a threshold beyond which they lose their adversarial property (example- if an image is cropped beyond a few pixels). As a result, adversarial images fail to fool object detection, instance segmentation, and image captioning models. We also found that when an adversarial image is generated sequentially across two networks, it may be able to fool the networks simultaneously.

Our work shows due to the lack of transferability and scalability of adversarial examples, they are yet to become a threat in the real world and are currently only effective in controlled experiments.

\section{Acknowledgement}

The authors would like to thank the anonymous reviewer(s) for their valuable comments and
suggestions and the University of Illinois at Urbana Champaign for their GPU computing without whose support the paper would not have been possible.

{\small
\bibliographystyle{ieee}
\bibliography{egbib}
}

\end{document}